\documentclass[journal]{vgtc}                

\usepackage{mathptmx}
\usepackage{graphicx}
\usepackage{times}
\usepackage[nocompress]{cite}
\usepackage{todonotes}

\usepackage[bookmarks,backref=true,linkcolor=black]{hyperref} 
\hypersetup{
  pdfauthor = {},
  pdftitle = {},
  pdfsubject = {},
  pdfkeywords = {},
  colorlinks=true,
  linkcolor= black,
  citecolor= black,
  pageanchor=true,
  urlcolor = black,
  plainpages = false,
  linktocpage
}

\onlineid{0}

\vgtccategory{Research}

\vgtcinsertpkg

\title{In Automation We Trust\\Investigating the Role of Uncertainty in Active Learning Systems}

\author{Michael L. Iuzzolino, Tetsumichi Umada, Nisar Ahmed, Danielle Albers Szafir}
\authorfooter{
\item
Michael L. Iuzzolino is with University of Colorado Boulder. E-mail: michael.iuzzolino@colorado.edu.

\item
Tetsumichi Umada is with University of Colorado Boulder. E-mail: tetsumichi.umada@colorado.edu.
 
\item
Nisar R. Ahmed is with University of Colorado Boulder. E-mail: nisar.ahmed@colorado.edu.

\item
Danielle A. Szafir is with University of Colorado Boulder. E-mail: danielle.szafir@colorado.edu.
}

\shortauthortitle{Biv \MakeLowercase{\textit{et al.}}: In Automation We Trust}

\abstract{
We investigate how different active learning (AL) query policies coupled with classification uncertainty visualizations affect analyst trust in automated classification systems. A current standard policy for AL is to query the oracle (e.g., the analyst) to refine labels for datapoints where the classifier has the highest uncertainty. This is an optimal policy for the automation system as it yields maximal information gain. However, model-centric policies neglect the effects of this uncertainty on the human component of the system and the consequent manner in which the human will interact with the system post-training. In this paper, we present an empirical study evaluating how AL query policies and visualizations lending transparency to classification influence trust in automated classification of image data. We found that query policy significantly influences an analyst's trust in an image classification system, and we use these results to propose a set of oracle query policies and visualizations for use during AL training phases that can influence analyst trust in classification.
} 


\CCScatlist{ 
 \CCScat{K.6.1}{Management of Computing and Information Systems}%
{Project and People Management}{Life Cycle};
 \CCScat{K.7.m}{The Computing Profession}{Miscellaneous}{Ethics}
}

  \teaser{
 \centering
 \includegraphics[width=\textwidth]{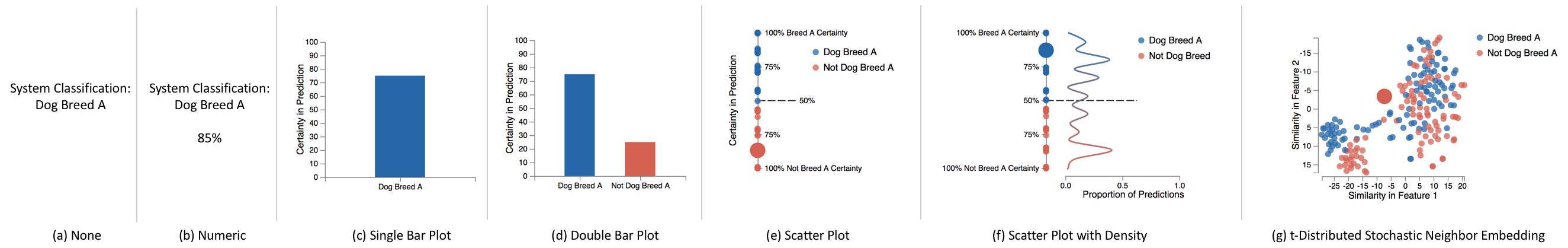}
  \caption{
  	We evaluate the role of classification uncertainty on analyst trust in active labeling systems. We visualize classification uncertainties using seven different techniques (a)-(e) designed to reveal increasing amounts of information about an algorithm's classification decisions and pair these visualizations with five different query policies. Our results suggest that active labeling system query policies and uncertainty visualizations can significantly influence analyst trust in the resulting classification. 
  }
  \label{figure:visualizations}
  }

\begin{document}



\maketitle

\section{Introduction}
  Automated systems are transforming all domains of human activity, from social media \cite{zeng2010social, wailthare2018artificial, anderson2017systems} to military intelligence \cite{cummings2004automation, cummings2007automation, pereira2006system}. Many such systems leverage machine learning to infer patterns from large collections of data. However, the datasets these systems use often contain large numbers of unlabeled datapoints or are insufficient to capture nuanced elements of expert knowledge. Increasingly, systems can address these limitations by integrating human analysts into the classifier training process using techniques like active learning. Visual-Interactive Labeling (VIL) accomplishes this by querying oracles with pre-labeled images to confirm the correctness of the label, whereas other approaches may query the oracle for a classification without any a priori label. Active learning is a form of semi-supervised machine learning that incorporates an analyst's knowledge into the algorithm's training, typically for classification tasks. The oracle (human analyst) is queried (asked by the system) to provide knowledge in the form of labeling instances the system is most uncertain of. Consequently, the algorithm is capable of achieving higher accuracies, particularly for datasets for which labeled data is sparse. Active learning is especially important for modern machine learning contexts where copious amounts of data are readily available, but the majority is unlabeled due to the difficult, time-consuming, and expensive nature of hand-labeling immense datasets. 

 A focus of active learning has been on optimal candidate selection strategies, where the best strategies are those that optimize the learning of the algorithm. These strategies may identify candidate data points selected from large collections of unlabeled data that yield optimal information gain. However, these approaches focus exclusively on best strategies for the automation system and do not account for effects on the analyst and their confidence in the system. For example, presenting analysts with only low confidence examples may bias analysts' impressions of the system's capabilities, priming analysts to mistrust the system \cite{atoyan2006trust, wilson1996new}. Alternative query policies may exist that yield sufficient algorithmic information gain without compromising the oracle's trust in the classification system post-training. In this study, we investigate how these query policies and the methods used to communicate classifier confidence during active learning might prime analyst trust in automated classifications.

  Trust in classification is an increasingly critical component of active learning applications as the analyst's interaction with the system does not stop at training: a growing breadth of applications relies on automated systems to inform decision making. Analysts are increasingly required to make decisions with data processed by autonomous, yet the query methods used may have unintended priming effects that may lead to mistrust or over-reliance. Machine learning visualization has been leveraged to increase analyst understanding of classifications through offering varying degrees of model transparency. However, conveying intuitive understandings of complex models and automation processes is challenging and many basic questions surrounding cognitive bias and human-computer interaction remain unanswered. Cultivating an understanding of the effects of various query policies and system uncertainty visualizations is paramount for building optimal interactions between humans and machines. 
  
 In this paper, we explore various query policies and their impact on oracle trust in a classification system by investigating how oracles may be affected by different query policies and visualizations of system classification uncertainty. We explore this question in a controlled experiment measuring how the query policy and uncertainty information available during active learning influence analyst agreement with and trust in autonomous classifications made by the algorithm.

 We created five AL query policies coupled with seven visualizations of the classifier's uncertainty and explored the effects of these combinations on analysts' trust in the classification of the system post-training. We found that policies restricted to high system confidence result in a significantly lower number of label changes and overall higher trust in the system, both during evaluation and post-evaluation. Additionally, significant interaction effects were discovered between query policies and visualizations, indicating that the visualization of classification uncertainty influences analysts' label-flipping behaviors as well as their perceived trust in the system during evaluation.
   
  \noindent\textbf{Contributions:}
  The main contribution of our work is to provide a set of empirical results measuring how query policies and visualization influence analyst trust in autonomous systems using active learning. These results allow us to identify candidate policies and representations that can tailor analyst trust to the needs of the target application.  
  
  \begin{itemize}
    \item We show that policies that only present images of high system confidence yield significantly lower label changes
    \item Policies that progress from low confidence image classification to high confidence image classification, and vice-versa, may result in anchoring effects that influence the analyst's trust in the system  
    \item When comparing the effects of query policies and visualizations on image classification tasks in the active learning paradigm, the predominant factor in affecting analyst trust in a system may be the query policy.
  \end{itemize}
  
\section{Related Work}
Recent work in visual analytics has explored how visualization can support analysts in more effectively leveraging machine learning for data analysis. For example, visualization can communicate aspects of a classifier's internal state \cite{wongsuphasawat2018visualizing,strobelt2018lstmvis}, classification performance \cite{talbot2009ensemblematrix,sarikaya2014visualizing,krause2016interacting}, and aid in model debugging \cite{liu2018visual,krause2017workflow}. While these approaches help bridge human analysts and automated classification, we have little empirical understanding of how the interaction between people and classifiers may influence analysts' use of classification outputs. To explore the role of transparency and human-in-the-loop approaches on analyst confidence, we draw on prior work in machine learning, psychology, and visualization.
In this section, we survey prior results from active learning, automation bias, and uncertainty visualization techniques relevant to our study.

\label{section:related_work}

\subsection{Active Learning and Visual-Interactive Labeling}
As datasets increase in size and complexity, active learning can help attenuate the exhaustive efforts required for generating labeled datasets by querying analysts to provide relevant information for important subsets of images. Work in this field has focused on devising model-centric approaches that optimize information gain from the fewest number of instances possible by querying an oracle with according to a prescribed query policy (see Fu et al. \cite{fu2013survey} for a survey). For example, the system may ask analysts to label datapoints the classifier is least certain about \cite{lewis1994sequential,holub2008entropy} or candidate datapoints representative of a cluster of similar instances \cite{zhou2009multi}. While these approaches help maximize the information gained from each labeled example, active learning does not provide analysts with a means for influencing the selection of candidate instances. Moreover, active learning proves particularly challenging when initiating the learning with no labeled instances (i.e., the bootstrapping problem \cite{mccallumzy1998employing}), and does not scale well for large datasets when asking analysts for single or multiple labels \cite{bernard2018comparing}. 

Visual-Interactive Labeling interfaces allow analysts to interactively explore the dataset and identify instances or groups of instances of interest, typically via interactive visual interfaces \cite{seifert2010user, hoferlin2012inter, bernard2017unified, bernard2018comparing}. VIL techniques are user-centric; the analyst leverages their domain knowledge to find useful instances rather than relying heavily on algorithmic sorting and filtering of the dataset. Human analysts have tremendous pattern recognition capacities; Roads and Mozer have shown that even untrained novices can accurately make difficult classification decisions \cite{roads2017improving}. While VIL enables the analysts to directly employ domain knowledge and superior high-level pattern recognition capabilities that automation systems generally lack, the analysts' selection of instances carries bias, thereby having the potential to significantly decrease the performance of the model. 

An optimal solution for generating labeled data would be one that leverages the strengths of both VIL and active learning \cite{bernard2018comparing}. Although previous work has suggested the integration of the techniques, no work had been done on directly comparing the strategies \cite{heimerl2012visual, hoferlin2012inter, seifert2010user, settles2011closing}. Bernard et al. addressed this gap by conducting an experimental study that directly compared the performance of active learning labeling strategies against the those employed by VIL. They examined how labeling task complexity impacts the different strategies and explored the differences between single- and multi-instance labeling scenarios \cite{bernard2018comparing} and found that while active learning outperforms VIL in terms of effectiveness, VIL leads to a significant increase in classification efficiency and is particularly better at simple tasks and for bootstrapping. Despite these promising findings, these model-centric approaches do not explore the effects of labeling policy on analyst trust. 

Recent work has explored model- versus human-centric modeling approaches. Tam et al. provide an in-depth analysis of machine learning versus visual analytics approaches to classification model building and found that the human-centric approaches provide significant improvements to the model building task, especially under the condition of sparse datasets \cite{tam2017analysis}. They argue that the incorporation of an analysts' ``soft knowledge" fosters improved performance and concluded that a combination of machine learning and visual analytics, where the strengths of each approach are leveraged, will provide the best performance. Our work in this paper will expand upon this notion by exploring the counterbalancing of machine- and analyst-centric optimizations to better understand the role of instance selection on analyst trust.

\subsection{Automation Bias}
While AL methods hope to improve classification for sparely labeled data, people ultimately rely on the resulting classifiers to autonomously process large quantities of data to aid identification and decision making. Analyst trust in a system can significantly influence their reliance on classifier outputs. For example, automation bias is the tendency for operators to under- or over-rely on automation. This topic has been heavily studied in domains such as healthcare \cite{goddard2011automation_healthcare, coiera2006safety, goddard2014automation} and aviation \cite{mosier1998automation, skitka1999does, parasuraman1997humans, parasuraman2010complacency} due to the critical outcomes of sensitive decision making processes in these domains. The tendency for operators to under-rely on automation may have resulted in the Costa Concordia cruise ship disaster that killed 32 passengers after the ship's captain diverted from the course set by the automation \cite{hoff2015trust}, whereas over-reliance on automation may have contributed to the Turkish Airlines Flight 1951 crash in 2009 after the altitude-measuring instrument failed. The crash killed nine people and was believed to be partially caused by the pilots' over-reliance on the autopilot \cite{hoff2015trust, de2014duration}. 

In medicine, clinical decision support systems (CDSS) undergird decisions that directly affect patient outcomes. For example, a CDSS developed for cervical cancer screening achieves recommendation accuracies of 93\%, a substantial improvement over the previous 84\% \cite{ravikumar2018improving}. Although modern decision support systems provide ever-increasing performances, they still exhibit errors. Importantly, these errors have been demonstrated to reverse an analyst's correct assessment and are representative of automation bias and complacency and result from either an attentional bias or insufficient attention and monitoring of automation system output \cite{goddard2011automation}. Parasuraman et al. show that these types of errors represent, ``different manifestations of overlapping automation-induced phenomena, with attention playing a central role" \cite{parasuraman2010complacency}. 

Trust is tightly coupled with automation bias, and a considerable amount of research on trust has been conducted in the fields of psychology, sociology, philosophy, political science, economics, and human factors. Trust can be viewed as a belief or attitude \cite{rotter1967new, heineman_1984, rubin1994social}, but others view it as a willingness to accept vulnerability \cite{mayer1995integrative, deutsch1960effect}. In human factors research, a focus has been placed on the role of trust in guiding interactions with various technologies \cite{hoff2015trust}. Corritore et al. developed a model of online trust between users and websites and identified three mediating factors: perception of credibility, ease of use, and risk \cite{corritore2003line}.   

Research on automation trust shows that analysts will adjust their trust towards automation decision aids to varying degrees as a function of valid or invalid system recommendations \cite{yang2016users}. Specifically, Yang et al. demonstrated that the magnitude of trust decrement is greater than that of trust increment for invalid and valid system recommendations. Their work also addressed contrast effects: the tendency of a decision maker to make evaluations of a system by benchmarking against other systems, individuals, or the self. That is, the automated system is evaluated favorably when the analyst is less capable of completing the task, but more harshly when the analyst is capable and the system makes an invalid recommendation. 

\subsection{Uncertainty Visualization}
As imperfect automation systems integrate deeper into decision-making roles with safety-critical outcomes, such as medicine and military, developing means for effectively communicating the system's uncertainty over its output becomes paramount for the success of the system and the well-being of the human operators and bystanders. Uncertainty information has a rich and growing history of research \cite{pang1997approaches, bonneau2014overview, potter2012quantification, maceachren2012visual}, but it is classically difficult to measure, understand, and visualize \cite{pang1997approaches}. 

The methods used to visualize uncertain data can significantly alter analysts' abilities to effectively reason under uncertainty \cite{correll2014error}. For example, Potter et al. reexamine the canonical box plot and develop a new hybrid summary plot that incorporates a collection of descriptive statistics, such as higher order moments (skew, kurtosis, excess kurtosis, etc.), to highlight salient features of the data \cite{potter2010visualizing}. 

Padilla et al. explored the influence of different graphical displays on non-expert decision making under uncertainty \cite{padilla2015influence}, testing whether different glyphs and overall level of variability displayed would influence decisions about the accuracy of a weather predictor. The study revealed both the underlying variability of the data distribution and the visualization of that uncertainty affected the participant's decisions. We build on these findings to explore how different methods of communicating uncertainty associated with classification outcomes during active learning influence analyst trust in automated classification.

\subsection{Priming and Anchoring Effects}
Priming and anchoring effects have a rich history of investigation in human psychology research \cite{wilson1996new, strack1997explaining, jacowitz1995measures}. Priming is the phenomenon in which human decision making is influenced by a preceding perceptual stimulus \cite{wilkinson2005graph} whereas anchoring effects describe a type of cognitive bias that describes a tendency for humans to rely too heavily on a single piece of information---typically, the first instance of information provided---to guide their decisions. Ergo, this piece of information acts as a conceptual anchor. 

Recent work shows that these effects play a role in perceptual tasks within the domain of visualization. For example, Valdez et al. utilize visual class separability in scatter plots as a perceptual task in visualization. They show that humans judge class separability of the same visualization differently depending on exposure to previously seen scatter plots \cite{valdez2018priming}. In our work, we draw upon this research to explore priming and anchoring effects in the context of active learning query policies.


\section{Motivation \& Research Questions}
In this study, we seek to address three research questions relevant to active learning systems: 
\begin{enumerate}
	\item [\textbf{RQ$_1$}] Will query policies formulated on system confidence affect the analyst's confidence in the system during autonomous classification?
	\item [\textbf{RQ$_2$}] How will visualizations of classifier confidence provided during active labeling influence trust in autonomous classification? 
	\item [\textbf{RQ$_3$}] Do visualizations influence the effects of query policies on trust?
\end{enumerate}

We address these questions in two phases. In the first phase, we conduct a preliminary data collection study to formulate a ground truth confidence based on people's performance on an image classification task. In the second phase, we use these confidences to simulate a synthetic classifier aligned with anticipated human confidence in classification. We ask participants to actively train the classifier based on a set of queried examples selected using one of five tested query policies paired with seven visualization types the provide increasing amounts of information about classifier confidence regarding the queried image. We then gauge participant trust in the resulting algorithm using a series of objective (percentage agreement) and subjective (self-reported perceptions of deception, trust, and collaboration) metrics.

\subsection{Query Policies}
In active learning, an algorithm determines the instances from an unlabeled datasets that will yield an optimal information return from querying an oracle for the correct label. However, in this model-centric approach, the effects on the analyst's confidence in the system post-training is not accounted for. Drawing on prior research in policies related to learning \cite{carvalho2014effects, carvalho2015benefits, carvalho2017sequence} and psychology \cite{yang2016users, valdez2018priming}, we devised five query policies to investigate the potential effects of query policy on analyst confidence summarized in Table \ref{table:policies}. 

The first policy randomly samples images from the full dataset, resulting in a sequence of images with a random distribution of confidences. The work of Carvalho \& Goldstone on learning policies indicate that interspersed versus blocking policies influence learning capabilities \cite{carvalho2017sequence}. Consequently, we can explore these interspersed (RAND) versus blocking effects (ALC, AHC) by developing policies that randomly sample images that are exclusively low or high confidences, respectively. The last two policies, LtH and HtL, randomly sample images from the set and then arrange them from low to high and high to low, respectively. The work on priming and anchoring effects suggests that the first instances of learning may overwhelmingly govern the decision-making tasks in an evaluation phase, despite the information provided at the trailing end of a learning phase \cite{valdez2018priming}. These two policies allow us to explore these effects.

During the training phase, the analyst is sequentially presented with 25 images and provided the system's classification for the given image. Each classification label has a corresponding confidence and correctness (detailed in \S \ref{section:research_study}). While prior work has explored the effects of correctness on trust in automated systems \cite{yang2016users}, we instead elect to focus on confidence as active learning query policies seldom have a known ground truth to compute correctness for query instances. Different policies correspond to the ordering of the images based on confidence. We conducted a crowd-sourced study of human confidence on our tested image corpus to compute a ground truth confidence for our simulated classification system in order to align our query policies with participant perceptions.

We then thresholded these scores based on the confidence distribution of the full image corpus to identify high- and low-confidence images to seed our corpus. More details on policy implementation are discussed in Sections \ref{section:data_collection_study} and \ref{s2_stimuli}.

\begin{table}
\caption{Query Policies} 
\label{table:policies}
\centering 
\begin{tabular*}{\columnwidth}{c c c c c c}
\hline\hline
& & Policy & & & Description\\
\hline
& & RAND & & & Confidence randomly interspersed \\
& & ALC & & & All low confidence \\
& & AHC & & & All high confidence \\
& & LtH & & & Low to high confidence \\
& & HtL & & & High to low confidence \\
\hline
\end{tabular*}
\end{table}

\subsection{Visualizations}
We developed seven conditions for visualization (see Table \ref{figure:visualizations}) that span a spectrum of classification transparency. The first visualization provides no information about classification uncertainty and therefore no transparency into the system's certainty or decision-making processes. The numeric visualization provides a raw confidence score, offering a rudimentary level of transparency into the systems' classification certainty. The two bar charts visualize this same uncertainty score in an alternative manner. Next, the scatter plot visualizations provide a higher level of transparency by offering the analyst contextual information about the current image's classification and degree of certainty within the population of images. Lastly, t-SNE, an approach commonly used in modern visualization tools \cite{hung2017making, maaten2008visualizing, bernard2018comparing, turkay2017challenges} visualizes the population of images in an embedded feature space, again displaying the current image, its classification and degree of certainty within the larger population of dataset images. 

\section{Data Collection Study}
\label{section:data_collection_study}

We first conducted a data collection study aimed at generating an empirical measurement of human classification uncertainty for use in the primary research study. We utilized crowd-sourced measurements to simulate classification in our primary study as opposed to a machine learning classifier in order to mirror participant's certainty in the classification and to avoid classifier-dependent results. We conducted a 6 (image class) x 4 (downsampling rate) mixed factors experiment measuring human accuracy at an image classification task to establish this ground-truth confidence dataset.

\subsection{Stimuli}
\label{s1_stimuli}
\begin{figure}
 \centering
 \includegraphics[width=\columnwidth]{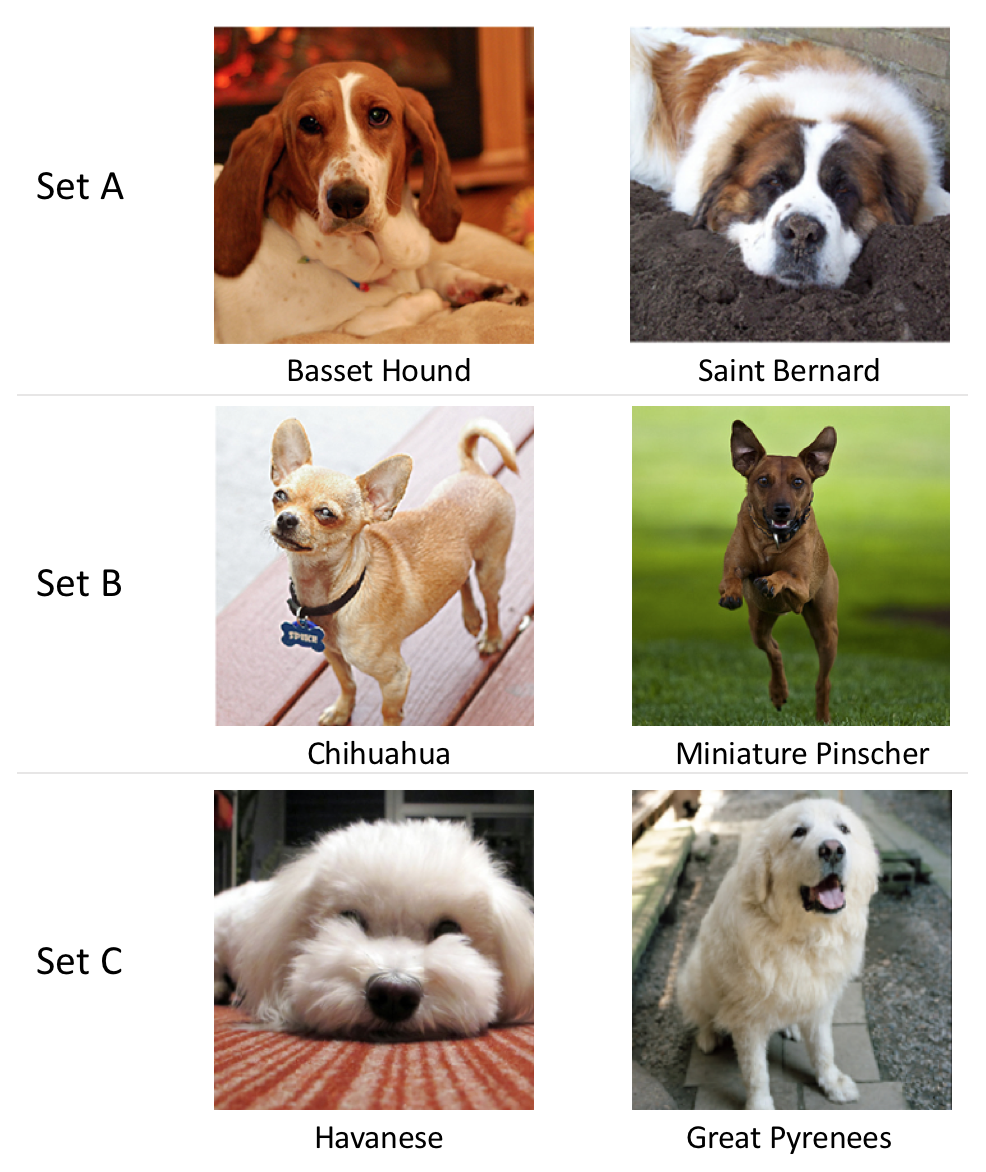}
  \caption{Representative images of dog breeds utilized to construct the datasets used in our study.}
  \label{figure:dataset_examples}
\end{figure}

We selected our stimuli from a collection of dog breed images drawn from the Oxford-IIIT Pet Dataset \cite{parkhi2012oxford}---a 37 category pet dataset with roughly 200 images for each category (see Figure \ref{figure:dataset_examples}).

Two categories of dog breeds were chosen heuristically for each set such that they shared a sufficiently large subset of common features such that they may prove challenging to discriminate on during a classification task. Twenty-five images from each of the six dog breeds were sampled. From each of the 25 images, 4 new images were generated via down-sampling using maxpooling operations at four different down-sampling rates: 2, 4, 7, and 10. For example, for a down-sampling rate of 2, 2x2 pixel grids would be pooled into a 1x1 pixel grid for the next image, where the 1x1 pixel is the maximum pixel value from the 2x2 pixel grid of the original, full-resolution image. This generated a dataset of 2,400 images (3 breed pairs $\times$ 200 images per breed pair $\times$ 4 downsampling rates).

\subsection{Experimental Design}
Our independent variables were dog breed and downsampling rate. Dog breed consisted of six levels, one for each breed (Basset Hound, Saint Bernard, Chihuahua, Miniature Pinscher, Great Pyrenees, and Havanese), tested as a between-subjects factor, with each participant asked to classify whether or not an image is a member of one breed. Distractor breeds for each primary breed are shown in Figure \ref{figure:dataset_examples}. Down-sampling rate consisted of four levels (2, 4, 7, and 10), tested as a within-subjects factor. Our dependent variable is accuracy, which is measured as the proportion of correct labels applied by the population.

\subsection{Experimental Task}
Our study consistent of four phases: (1) informed consent, (2) training, (3) labeling, and (4) demographics. In phase one, the participant provides informed consent in accordance with our IRB protocol. After accepting the terms, the participant proceeds to phase 2, the tutorial for the study. In the tutorial, the participant is presented with an example image for each of the two dog breeds at the top of the screen to aid their training, a set of instructions, and a single test image of either the tested breed ("dog breed A") or its complement within the tested set ("not dog breed A"). Note, actual dog breed names are not used to provide consistency across datasets. The test image is presented in the middle of the screen, and the participant is instructed to label it as either \textit{dog breed A} or \textit{not dog breed A} by pressing either key \textit{F} (\textit{dog breed A}) or \textit{J} (\textit{not dog breed A}). These keys align with natural pointer finger positioning on a QWERTY keyboard.

We draw 50 non-downsampled images randomly from each breed pair (25 per breed) with images counterbalanced between participants to ensure an equal distribution of responses. The images in the tutorial segment are randomly selected without replacement from this set, resulting in 25 tutorial images. The participant must correctly label 5 images from each breed before moving onto phase 3, indicating they have sufficiently learned to discriminate between the two breeds. The remaining 25 images are used in phase 3.

In phase 3 the participant is no longer provided with examples of the dog breeds. They are instead informed them to label the currently displayed image as either \textit{dog breed A} or \textit{not dog breed A} using the same keypress inputs as in the tutorial. Participants complete this identification task for each of the remaining 25 images at each of the four tested downsampling rates, with images presented in random order to mitigate potential transfer effects. Four full resolution images are inserted at trial numbers 20, 40, 60, and 80 to serve as attention checks, resulting in the participant labeling a total of 104 images.

Lastly, in phase 4 the participant is asked to fill out demographic information.

\subsection{Participants}
178 participants with an average age of 37.7 ($\sigma=12.6$) and gender distribution of 85 males and 91 females, and 2 no replies participated in the experiment via Amazon Mechanical Turk. A total of 38 participants were excluded from the study resulting in 120 valid participants (10 measures per image $\times$ downsampling rate). Participant data was excluded from the study if the participant labeled all images with the same label or if they failed four attention checks.

\subsection{Results}
\begin{figure}
\centering
 \includegraphics[width=\columnwidth]{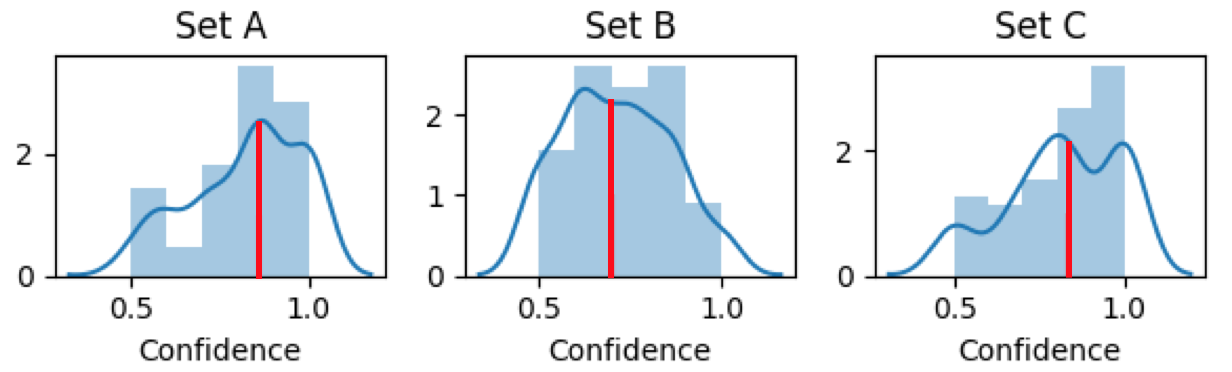}
  \caption{Summary of image classification confidence distribution across the three image datasets. Red line shows dataset median.}
  \label{figure:s1_dist}
\end{figure}

We computed confidence scores for each breed as the proportion of times each breed was labeled as ``Breed A'' for each of the six target breeds. We elected to use confidence rather than correctness as our primary metric to ensure our simulated classifier replicated the anticipated behaviors of the human oracle. Participants had a mean confidence for images in the breed pair for dataset A with 82\% on average ($\sigma=15\%$), dataset B with 71\% mean confidence ($\sigma=14\%$), and dataset C with 81\% mean confidence ($\sigma=19\%$). Figure \ref{figure:s1_dist} summarizes the distribution of classifications across the three image datasets.

The performance distribution within these datasets allows us to draw from a broad dataset of low- and high-confidence images for each breed pair in order to apply different query policies to the training phases of our primary experiment. We use the statistical mean of each dataset to divide our corpus into low- and high-confidence scores. This division created 104, 108, and 99 low-confidence images and 96, 92, and 101 high-confidence images for datasets A, B, and C, respectively.

\section{Primary Study}
\label{section:research_study}
We use the results of our data collection study to simulate an active learning classifier leveraging different query policies and classification uncertainty visualizations in order to understand the role of query policy and visualization on analyst trust in a system trained using active learning. We conducted this analysis using a 5 (query policy) $\times$ 7 (visualization method) mixed factors experiment conducted on Mechanical Turk.

\subsection{Stimuli}
\label{s2_stimuli}
Our study again used the three breed pair image datasets discussed in \S \ref{s1_stimuli}.
We used the ground truth accuracy from our data collection study (\S \ref{section:data_collection_study}) to generate the stimuli for our second experiment, with high confidence defined as 79\% or greater human classification accuracy and lower than 79\% defined as low classification accuracy. We implemented each of the five tested policies by sampling images within each breed based on these thresholds.

The seven between-participant visualizations used in the study are provided in Figure \ref{figure:visualizations}. The data generated from \S \ref{section:data_collection_study} was utilized to generate the visualizations. The color channel was used to represent the dog breed. In all visualizations, blue marks correspond to the classification label \textit{dog breed A} and red corresponds to \textit{not dog breed A}. Classification confidence in the two scatterplot visualizations and the t-SNE visualization (Fig. 1e, 1f, and 1e), the tested dataset was represented as points in the scatterplot, with the current image being classified being represented by a significantly larger point size. Feature embeddings in the t-SNE plot were extracted using TensorFlow, an open-source machine learning framework for Python.

\subsection{Experimental Design}
\label{section:exp_1_design}
This experiment was designed as a two-stage 5 (query policy) $\times$ 7 (visualization type) full factorial between-participants experiment.

We tested two independent variables: query policy and visualization method. The query policy consisted of five levels: random (RAND, selected at random from all images), all low confidence (ALC, selected at random from the set of low-confidence images), all high confidence (AHC, selected at random from the set of high confidence images), low to high confidence (LtH, selected at random from all images and ordered by increasing confidence), and high to low confidence (HtL, selected at random from all images and ordered by decreasing confidence), summarized in Table \ref{table:policies}. Visualization methods consist of seven levels: 1) no visualization (classification label only), 2) numeric representation of confidence as a percentage, 3) a bar chart with a single bar representing confidence, 4) a bar chart with two bars where one bar represents confidence in \textit{dog A} and the other representing confidence in \textit{not dog A}, 5) a scatter plot along the y-axis depicting the entire 25 images and their corresponding level of label confidence, 6) a scatter plot along the y-axis depicting the entire 25 images and their corresponding level of label confidence with a corresponding probability density overlay, and 7) a t-Distributed Stochastic Neighbor Embedding (t-SNE) visualization \cite{maaten2008visualizing}. These visualizations are summarized in Figure \ref{figure:visualizations}.

We measured both objective and subjective dependent variables related to trust in classification. Our objective metrics consisted of label disagreement---the number of times an analyst flips the classification of a system-applied label---computed over a set of 25 images with a uniform distribution of high and low confidence and correctness randomly sampled from a binomial distribution weighted according to the confidence score of the image (e.g., if the image confidence is 75\%, the system will provide the correct label in 75\% of instances). Percentage agreement is a meaningful dependent measure of trust; the more an analyst changes a label, especially when levels of self-certainty are high, the less the analyst is willing to rely upon the system \cite{yang2016users}. Further, the correctness distribution allows us to randomize effects of correctness in a manner consistent with the anticipated behavior of a classifier. 

Subjective measures were taken per-image and on the full dataset in a post-survey questionnaire. Our per-image subjective measures consisted of three 7-point Likert-type questions (see Table \ref{table:per_image_subjective}). We used these responses to construct one scale (Cronbach's $\alpha>.65$) describing trust (3 items). The post-survey questionnaire subjective measures consisted of 13 7-point Likert-type questions augmented from Jian et al. \cite{jian2000foundations}. Irrelevant questions from this trust scale (e.g., "The system's actions will have a harmful or injurious outcome") were supplanted with more relevant questions to active learning, such as perceptions of the analysts' impacts on the system output. The full survey is available on the project website \url{https://github.com/CU-VisuaLab/Active-Learning-Query-Policies}. We constructed three scales ($\alpha=.65$) describing deception (6 items), trust (3 items), and collaboration (3 items).

We analyzed our objective measures and per-question subjective measures using a three-factor (visualization method, query policy, image confidence) full factorial ANCOVA with question ordering and self-reported experience with machine learning treated as random covariates. We analyzed scales constructed from our post-testing questionnaire using a two-factor (visualization type and query policy) full factorial ANCOVA with machine learning experience, and question ordering included as random covariates. We used Tukey's Honest Significant Difference test (HSD) for all post-hoc analysis ($\alpha=.05$).

\begin{table}
\caption{Per-Image Subjective Measure Questions}
\centering
\begin{tabular*}{\columnwidth}{c c}
\hline\hline

\# & Question\\
\hline
1. & How much do you agree or disagree with \\
& the system's classification? \\
2. & How confident are you in your decision \\
& to keep or change the system's classification? \\
3. & How confident are you in the system's classification? \\
\hline
\label{table:per_image_subjective}
\end{tabular*}
\end{table}

\subsection{Experimental Task}
The research study consists of five phases: (1) informed consent, (2) tutorial, (3) training, (4) testing, and (5) survey and demographics.

In phase 1, the participant is presented with a consent form. After accepting the terms, the participant proceeds to phase 2, which is the tutorial for the study. In the tutorial, the participant is presented with an example image for each of the two dog breeds at the top of the screen to aid their training, a set of instructions describing both the task and how to interpret the tested visualization, and a single image of either \textit{dog breed A} or \textit{not dog breed A} with an accompanying visualization (one of the seven possible visualizations). Note, as in the first study, actual dog breed names are not used. A single image is presented in the middle of the screen that the user must label as either \textit{dog breed A} or \textit{not dog breed A}. To label an image, the participant is instructed to press key \textit{F} to label the image as \textit{dog breed A} and to press key \textit{J} to label the image as \textit{not dog breed A}. The images in the tutorial segment come from the set of 50 images per breed generated for the tutorial set, as described in \S \ref{s1_stimuli}, and are randomly shuffled. The participant must correctly label 5 images from each breed before moving onto phase 3, indicating they have sufficiently learned to discriminate between the two breeds, as well as having familiarized with the accompanying visualization.

In phase 3, participants are instructed that they will be training a classifier to better discriminate between the two dog breeds the participant was trained on, simulating an active labeling protocol. The participant will be presented 25 images, one image at a time, along with the system's binary classification of that object---\textit{dog breed A} or \textit{not dog breed A} stemming from the confidence measure generated by sampling from a binomial distribution structured according to the confidence score for each image from the data collection study in \S \ref{section:data_collection_study}---and an accompanying visualization that may provide them with additional information for making their own decision about which dog breed the current image corresponds to. 25 training images are selected without replacement according to the query policy sampling described in \S \ref{s2_stimuli}. The participant is no longer provided with examples of the dog breeds.

Once the participant has finished training the classifier on the 25 images, they proceed to a holding page that instructs them that the classifier is updating based on the feedback the participant provided the system. While the system trains, which takes approximately 10 seconds with simulated progress indicated using a horizontal progress bar, the participant is then instructed that they will now be required to evaluate the system's performance, and they proceed to phase 4. In phase 4, the participant will be sequentially presented with another 25 images along with the system's binary classification of that object: \textit{dog breed A} or \textit{not dog breed A}, again stemming from the confidence measure generated from the first study and not truly updated by the participants training in the previous experiment phase.

The 25 images correspond to the test set images described in \S \ref{section:exp_1_design} with images randomly sampled and presented in a random order. Unlike in the previous segment the participant is not provided with an accompanying visualization. This enables us to focus exclusively on trust constructed during active labeling. Showing classification confidence post-labeling integrates confidence into the testing phase and could interfere with primed perceptions. Therefore, the participant must choose to accept the classifier's labeling of the given image as either correct or incorrect without the visualization. Additionally, for each image, the participant must fill out three 7-point Likert-type questions pertaining to their confidence in the system's classification, their confidence in their own classification, and how much they agree with the system's classification.

After completing the system evaluation stage, the participant proceeds to phase 5, where the participant fills out the post-survey questionnaire and a short demographics survey including self-reported experience with machine learning and data analytics tools.

\subsection{Participants}
For the second study, 630 participants with an average age of 34.8 (std = 11.7) and gender distribution of 284 males, 327 females, and 19 no replies participated in the experiment via Amazon Mechanical Turk. All participants were paid \$0.50 for completion of the study, which ranged from 5-10 minutes in duration. A total of 25 participants was excluded from the study resulting in 605 valid participants. Participant data was excluded from the study if all labels were inverted during the training phase or if their responses to \emph{How much do you agree or disagree with the system's classification?} did not align with their actual agreement for two or more images in the testing phase.

\section{Results}
We report results for the active learning query policy and visualization user experiment in three different parts. In \S \ref{results:query_policy_and_vis}, we report on the effects of query policy and visualizations on participants' propensity to change the system's classification label. In \S \ref{results:per_image_trust} we present the effects of query policies and visualizations on per-image scales describing participant confidence in each classification. Finally, in \S \ref{results:per_image_trust}, we report the findings of the effects of query policies and visualizations on the scales constructed from the full dataset.

\begin{figure}
 \centering
 \includegraphics[width=\columnwidth]{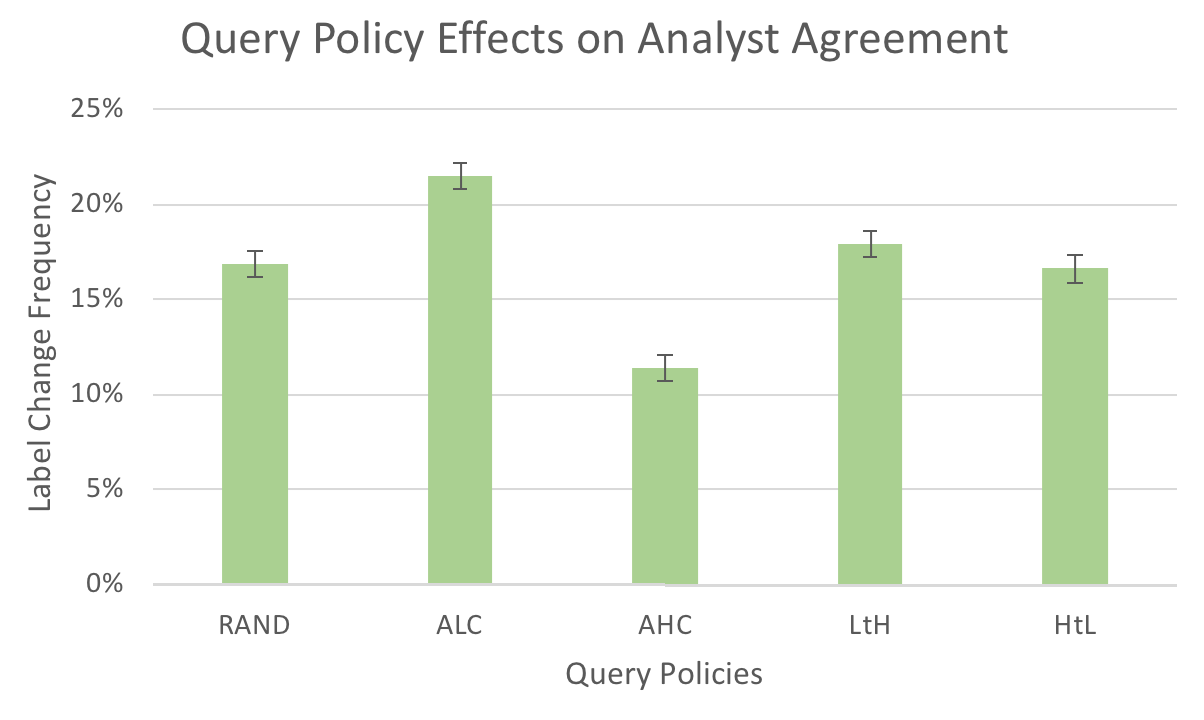}
  \caption{Comparison of query policy effects on analyst agreement with system classification. A significant difference exists between policies ALC and ALH, as well as between ALC and all other policies. These findings show that low system confidence results in participants disagreeing with system classification at a significantly higher frequency then with other query policies.}
  \label{figure:result_1}
\end{figure}

  \begin{figure}[b]
 \centering
 \includegraphics[width=\columnwidth]{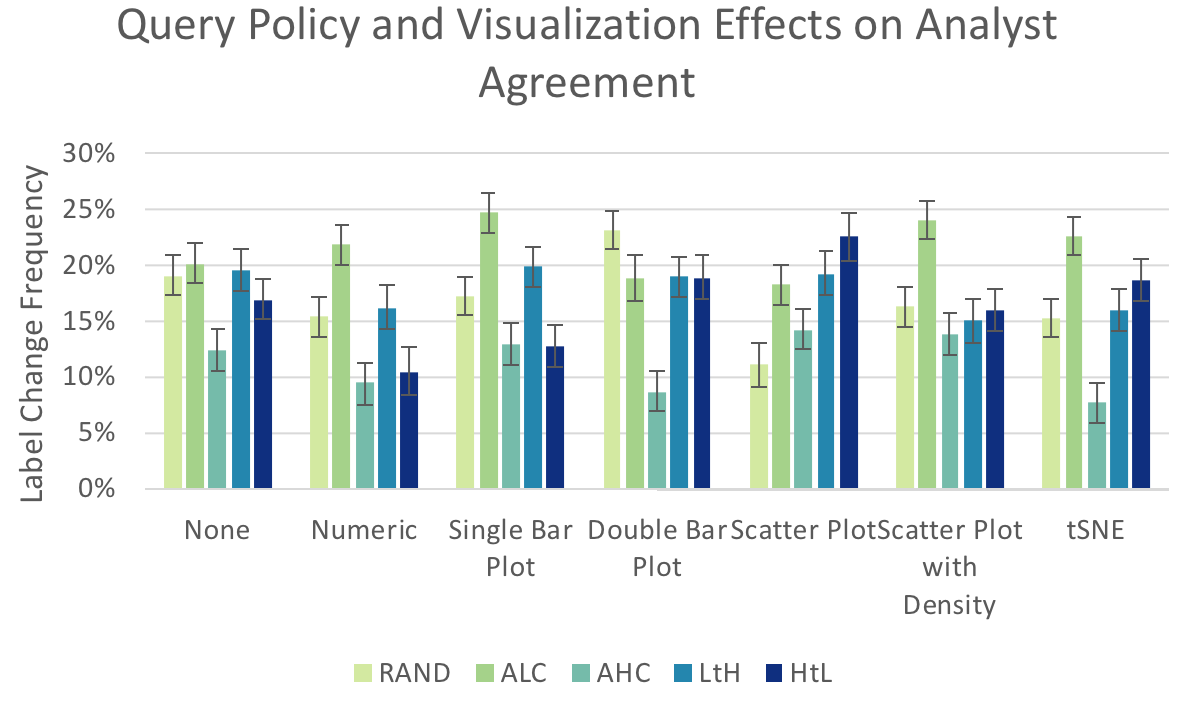}
  \caption{Comparison of query policy and visualization interaction effects on participants' agreement with system classification.}
  \label{figure:result_4}
\end{figure}

\subsection{Label Changing}
\label{results:query_policy_and_vis}
We analyzed the number of label changes---the measure of analyst agreement with the system's classification---of reported differences across experimental (query policy and visualization method) with machine learning experience and question number covariates and found a significant primary effect of query policy on analyst agreement (F(4,14180) = 27.6, $p < .0001$). Post-hoc analysis showed that participants change classification labels significantly more for ALC policies ($\mu=22\%, \sigma=0.69\%$) compared to AHC policies ($\mu=11\%, \sigma=0.70\%$). See Figure \ref{figure:result_1}. Participants also changed classification labels significantly more for ALC policies compared to LtH ($\mu=18\%, \sigma=0.72\%$), RAND ($\mu=17\%, \sigma=0.69\%$), and HtL ($\mu=17\%, \sigma=0.75\%$) policies. Further, AHC policies resulted in significantly fewer classification label changes than all other policies.

Significant secondary effects were found for interactions between query policies and visualizations (F(24,14180) = 2.8, $p < .0001$), query policy and classifier confidence (F(4,X) = 2.6, $p < .05$), visualizations and classifier confidence (F(6,14180) = 3.6, $p < .002$), as well as between query policy, visualization, and image confidence (F(24,14180) = 2.5, $p < .0001$). Post-hoc analysis shows that AHC query policies coupled with Numeric ($\mu=9\%, \sigma=1.9\%$), Single Bar Plot ($\mu=13\%, \sigma=1.8\%$), Double Bar Plot ($\mu=9\%, \sigma=1.8\%$), Scatter Plot with Density ($\mu=14\%, \sigma=1.9\%$), and tSNE ($\mu=8\%, \sigma=1.8\%$) visualizations yield significantly lower label-flipping as compared to ALC query policies coupled with either None ($\mu=20\%, \sigma=1.8\%$), Numeric ($\mu=22\%, \sigma=1.8\%$), Single Bar Plot ($\mu=25\%, \sigma=1.8\%$), Scatter Plot with Density ($\mu=24\%, \sigma=1.8\%$), or t-SNE ($\mu=23\%, \sigma=1.8\%$) visualizations. Additionally, AHC query policies coupled with any of the seven visualizations lead to significantly fewer label flips than an ALC query policy coupled with either Single Bar Plot or Scatter Plot with Density visualizations.

\subsection{System Trust: During Evaluation}
\label{results:per_image_trust}
We found a significant primary effect of query policy on classifier trust during evaluation (F(4,15040) = 72.3 $p < .0001$). Tukey's Test of HSD reveals that the AHC policy ($\mu=5.89, \sigma=0.029$) leads to significantly higher trust in the classifier for each image during evaluation compared to the RAND ($\mu=5.68, \sigma=0.028$), ALC ($\mu=5.23, \sigma=0.028$), LtH ($\mu=5.62, \sigma=0.029$), and HtL ($\mu=5.51, \sigma=0.030$) policies. In addition, RAND policies result in significantly higher system trust compared to HtL and ALC policies, but both LtH and HtL policies yield significantly higher trust than ALC policies. Significant secondary effects were found for interactions between query policies and visualizations (F(24,15040) = 6.2, $p < .0001$). Post-hoc analysis shows that AHC query policies coupled with any one of the seven visualizations provide significantly higher trust in the system as compared to ALC query policies coupled with either None, Scatter Plot with Density, or t-SNE visualizations.

\begin{figure}
 \centering
 \includegraphics[width=\columnwidth]{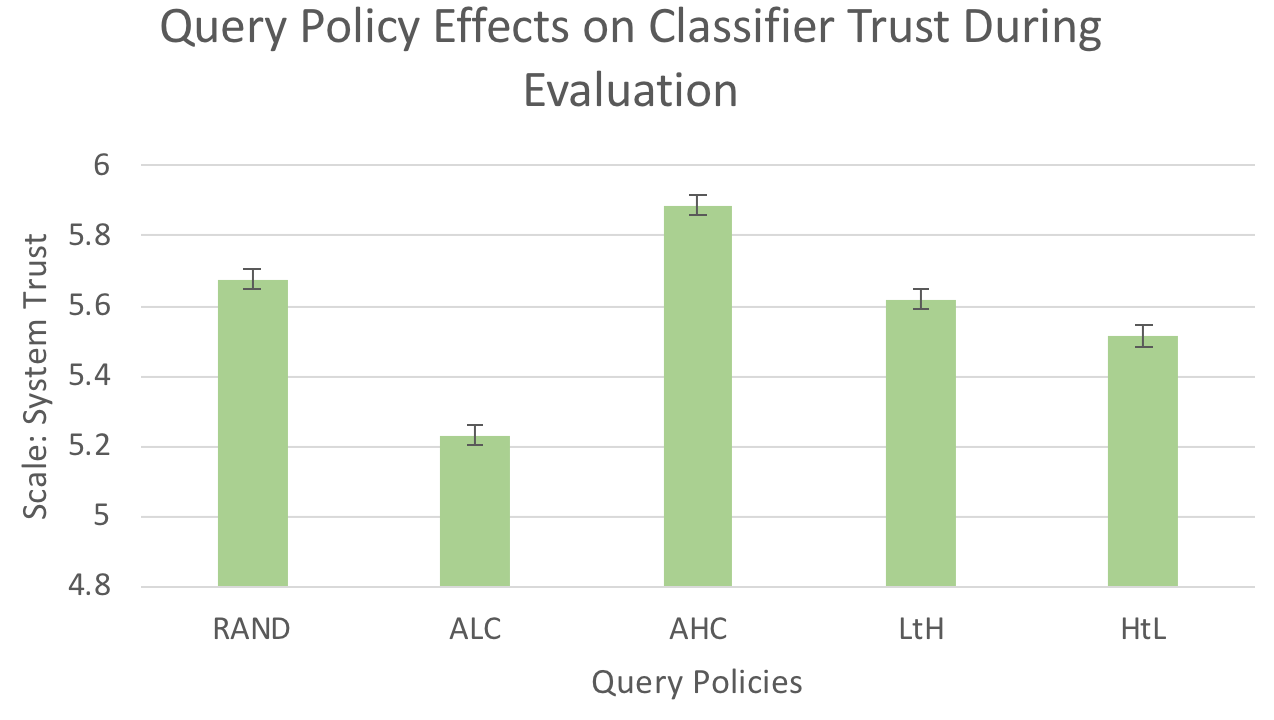}
  \caption{Comparison of query policy effects on classifier trust during evaluation. A significant difference exists between policies ALC and ALH, as well as between ALC and all other policies. These findings further coincide with our objective measure results, showing that low system confidence also influences participants to have lower trust in the system during evaluation.}
  \label{figure:result_2}
\end{figure}

\subsection{System Trust and Deception: Post Evaluation}
\label{results:dataset_trust}
We found significant primary effects of query policy on the participants' post-evaluation perceptions of the classifier as  deceptive (F(4,616) = 3.5 $p < .01$) and trustworthy (F(4,616) = 7.4 $p < .0001$). Tukey's Test of HSD reveals that the ALC policy ($\mu=3.38, \sigma=0.115$) significantly increases the participants' perceptions of system deception compared to AHC ($\mu=2.88, \sigma=0.120$) and HtL ($\mu=2.84, \sigma=0.111$) policies. Conversely, participants had significantly higher trust in the system for RAND ($\mu=4.50, \sigma=0.116$), AHC ($\mu=4.87, \sigma=0.116$), LtH ($\mu=4.46, \sigma=0.118$), and HtL ($\mu=4.55, \sigma=0.124$) policies compared to the the ALC policy ($\mu=3.98, \sigma=0.119$). No secondary effects across scales were observed, and no significant effects were found on the participants' perceptions of positive collaboration with the system.

\begin{figure}
 \centering
 \includegraphics[width=\columnwidth]{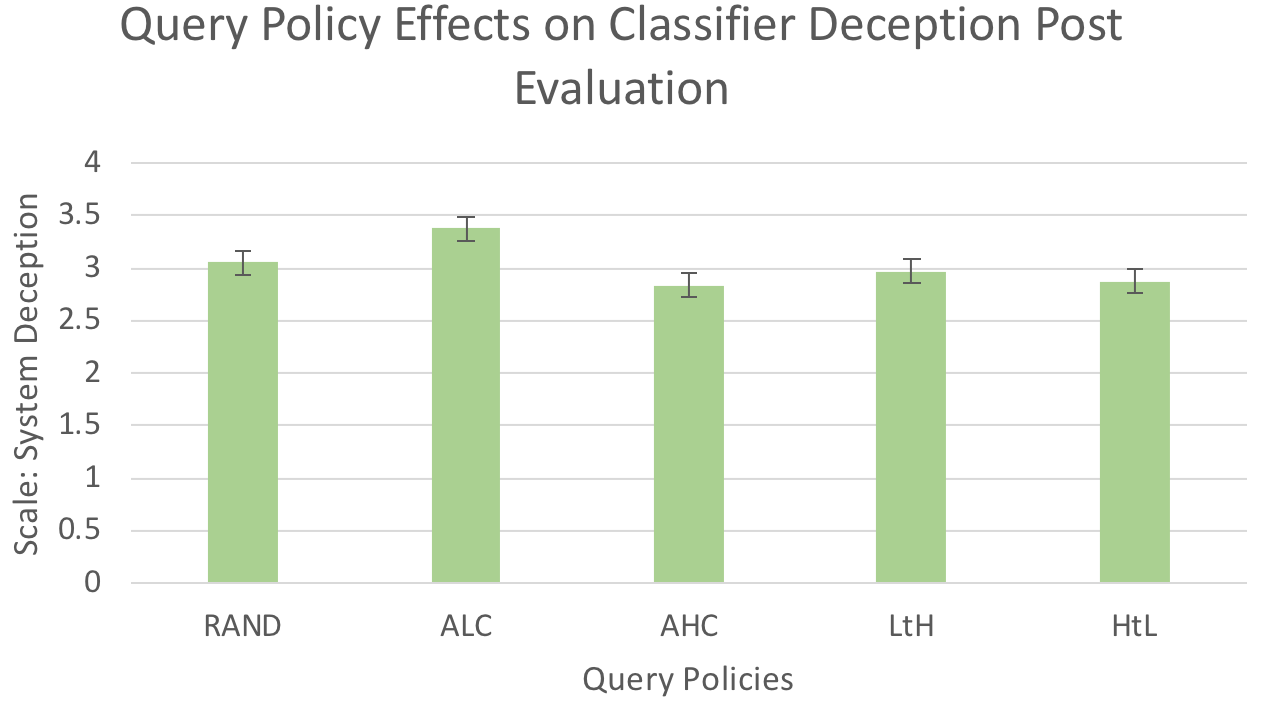}
  \caption{Comparison of query policy effects on participants' perceptions of classifier deception post-evaluation. A significant difference exists between policies ALC and AHC/HtL query policies. Low system confidence influences participants' overall perceptions of the system to be more deceptive.}
  \label{figure:result_3}
\end{figure}

\begin{figure}
 \centering
 \includegraphics[width=\columnwidth]{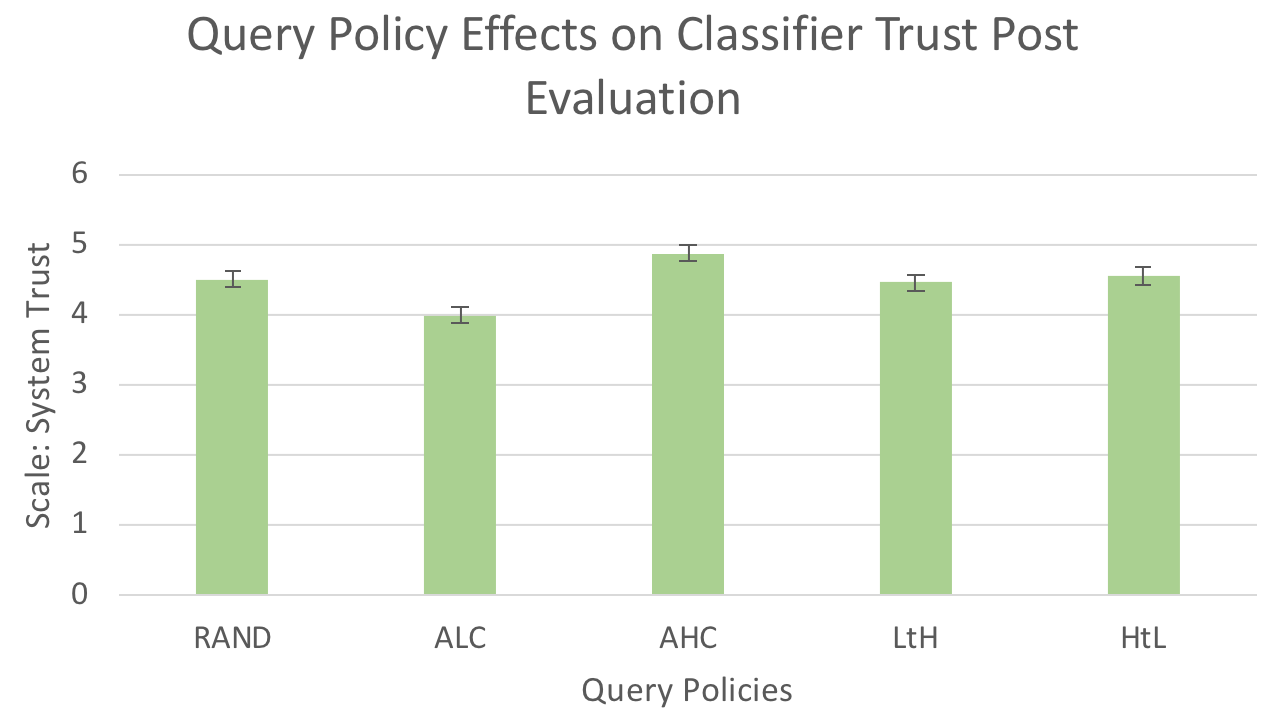}
  \caption{Comparison of query policy effects on participants' overall trust of the classifier post-evaluation. A significant difference exists between the AHC query policies and all other query policies. Systems that query with high classification confidence policies influence participants to be overall more trusting of the system's classifications.}
  \label{figure:result_4}
\end{figure}


\section{Discussion}
Our results indicate that query policy plays a significant role in modifying an analysts' objective behaviors and perceptions of the automated classification system. We have shown that query policies formulated on system confidence will affect the analyst's confidence in the system during autonomous classification. The analysts' label inversion behavior changed significantly when operating under the AHC query policy compared to all other query policies---particularly, the ALC policy. Interestingly, participants flipped labels significantly less for RAND, LtH, and HtL policies as compared to the ALC policy. Currently, the ALC query policy is used predominantly in active learning. Our results suggest that current active learning systems may be significantly affecting analysts' behavior with the classification system post-training.

The participants' perceptions of system trust during evaluation were significantly affected by the query policy. The AHC policy resulted in participants' having higher self-reported trust in the system compared to all other query policies. The results also demonstrate the HtL query policy yields significantly higher trust in the system during evaluation as compared to the ALC query policy. This may be suggestive of an anchoring effect: when analysts are first learning to use a system and the system immediately presents high confidence classifications, those instances anchor the analysts' perceptions toward having high confidence in that classifier measured by the participants' perceptions of trust during evaluation. One exception to this hypothesis is that the RAND query policy yields higher levels of analyst trust compared to the LtH and HtL query policies. Carvahlo \& Goldstone found evidence from category learning that interleaving policies like RAND cause people to attend more to differences in sequential items while blocked trials like LtH and HtL emphasize similarities between items \cite{carvalho2017sequence}. This emphasis on similarity may contribute to stronger anchoring effects, whereas RAND conditions may contribute to a stronger focus on the variety of queried images.

Similarly, the participants' perceptions of the system post-evaluation reveal that AHC query policies engender a sense of deception and lower levels of trust, whereas RAND, LtH, and HtL query policies again resulted in higher levels of trust compared to ALC. This finding, in conjunction with both findings for objective label-flipping behavior and trust during evaluation is important for the following reason. ALC query policies are the predominant policy for active learning systems because they are optimal for in terms of algorithmic information gain. However, we have demonstrated that the ALC query policy affects both objective behavior and subjective perceptions of the system post-training (\textbf{RQ$_1$}).

The influence of classification uncertainty visualization on the participants' behaviors and perceptions was only observed through secondary effects when coupled with specific querying policies (\textbf{RQ$_2$}, \textbf{RQ$_3$}). For instance, ALC query policies that attempt to visualize classification uncertainty with Single Bar Plot, Scatter Plot with Density, or t-SNE visualizations may increase an analysts' frequency of changing the system's classification label, and query policies coupled with Scatter Plot with Density visualizations may equally suffer in both objective performance and subjective perceptions. Although we suspect that future work may reveal deeper insights into the effects of visualization on active learning, ultimately, our results indicate that classification uncertainty visualization is not nearly as important as the query policy. This suggests that we must consider the cognitive interactions with active learning systems before we can meaningfully engage with different methods of introducing transparency.

Although AHC query policies had significant positive effects on both objective and subjective measures in all cases, this query policy is human-centric and would likely hamper the abilities of an active learning system. That is, if the active learning system only queries the oracle with data instances it is highly confidence about, it is effectively minimizing its information gain and thereby undermining the primary functionality of active learning systems. Our findings are important because they suggest alternative policies---RAND, LtH, or HtL---that may optimize both algorithmic efficiency and long-term trust in the system. This may be particularly important for domains such as healthcare and aviation, where analysts engage with automation systems over long periods of time. In these domains, cultivating and maintaining trust in such systems is critical for the optimal functionality and safety for the institution.

Overall, the results of query policies on the analysts' perceptions of the classifier post-evaluation indicate that policies consisting of exclusively low confidence classifications---the standard for current active learning systems---causes the analysts' to flip classification labels more frequency and have significantly lower trust in the system compared to all other query policy conditions. Moreover, our findings indicate that the ALC policy results in analysts' perceiving the system as more deceptive. Consequently, future active learning systems may significantly benefit from employing less model-centric querying policies in the case where long-term analyst trust is required. 


\subsection{Limitations \& Future Work}
Our study made a number of simplifying assumptions in order to investigate how interactions between analysts and active learning approaches shift analyst engagement with and trust in end classifications. For example, we focused on simple query policies that query an oracle for an image label for binary classification. Future directions that follow from our work may focus on more sophisticated query policies that explore priming and anchoring effects in greater detail and within specific use cases or scenarios. While we elect to provide a label alongside our query image to help analysts interpret presented confidence metrics, the effects of querying with and without an a priori system classification could provide further valuable information. Our simulated classification policies also may influence our results in practice: integrating these effects into actual machine learning algorithms is important future work for understanding how query policies may balance trust and performance across both people and classifiers. Future research avenues should also expand our binary classification work into multi-class classification scenarios to help determine the generalizability of the results in this paper. 

We additionally focus on visualizations exclusively of classification confidence rather than reflecting the internal state of the algorithm or how the algorithm's state changes as a result of analyst input. An expanded study exploring the role of internal state visualizations as visualization persistence in evaluating both training and testing behavior could inform visual analytics systems that enable continuous collaborative analyses between analysts and automated systems as well as continuous learning processes like Lifelong Learning \cite{silver2013lifelong}.

Additionally, we elected to use a simple binary classification task that requires no expert knowledge to complete. However, many applications of active learning and other human-in-the-loop approaches leverage analyst expertise to improve classification. Expert domains may, for example, introduce constraints around the maximal number of queries an algorithm can make or the risk tolerance for different application domains. Potential target domains include the exploration of alternative datasets, especially those in high-risk scenarios, such as medicine and missile defense.  


\section{Conclusion}
In this paper, we investigate how different active learning query policies coupled with classification uncertainty visualizations affect analyst trust in automated classification systems. We find that policies consisting of low confidence classifications significantly reduce trust in the automation system, as observed through both objective measures (number of classification label changes) and subjective measures on both individual images and over the full image set post evaluation. These policies also lead analysts to perceive the classification system as more deceptive. These results offer preliminary considerations for active learning systems to align analyst trust and behavior with the needs of the application domain. We envision these findings as first steps towards building more trustworthy active learning systems that focus on more reliable, long-term interactions between analyst and algorithm. 

\acknowledgments{
The authors wish to thank Steven Smart for his guidance on database management and Amazon Mechanical Turk insights. This work was supported in part by U.S. Air Force SMC-RSX \#FA8810-17-C-0006.}

\bibliographystyle{abbrv}
\bibliography{automation_query}
\end{document}